\documentclass[10pt,twocolumn,letterpaper]{article}

\usepackage{cvpr}              
\definecolor{cvprblue}{rgb}{0.21,0.49,0.74}
\usepackage[pagebackref,breaklinks,colorlinks,allcolors=cvprblue]{hyperref}
\usepackage{makecell}

\def\paperID{} 
\def\confName{CVPR}
\def\confYear{2026}

\title{Beyond Detection: Multi-Scale Hidden-Code for\\Natural Image Deepfake Recovery and Factual Retrieval}

\author{Yuan-Chih Chen\\
IIS, Academia Sinica, Taiwan, ROC \\
{\tt\small willpower057@iis.sinica.edu.tw}
\and
Chun-Shien Lu\\
IIS, Academia Sinica, Taiwan, ROC\\
{\tt\small lcs@iis.sinica.edu.tw}
}

\begin{document}
\maketitle
\begin{abstract}
Recent advances in image authenticity have primarily focused on deepfake detection and localization, leaving recovery of tampered contents for factual retrieval relatively underexplored.
We propose a unified hidden-code recovery framework that enables both retrieval and restoration from post-hoc and in-generation watermarking paradigms.
Our method encodes semantic and perceptual information into a compact hidden-code representation, refined through multi-scale vector quantization, and enhances contextual reasoning via conditional Transformer modules.
To enable systematic evaluation for natural images, we construct ImageNet-S, a benchmark that provides paired image–label factual retrieval tasks.
Extensive experiments on ImageNet-S demonstrate that our method exhibits promising retrieval and reconstruction performance while remaining fully compatible with diverse watermarking pipelines.
This framework establishes a foundation for general-purpose image recovery beyond detection and localization.
\end{abstract}    
\section{Introduction}
\label{sec:intro}

The rapid development of generative and editing tools has made digital image manipulation more accessible than ever.  
While this progress enables creative expression and efficient content generation, it also raises serious concerns regarding content authenticity and ownership protection.  
Recent advances in generative models such as Stable Diffusion~\cite{rombach2022StableDiffusion} have further blurred the boundary between real and synthetic images, making traditional watermarking and forensic detection techniques increasingly insufficient.  


In the literature, image deepfake detection and localization methods~\cite{xufakeshield,zhang2024editguard,huvideoshield} aim to identify or highlight modified regions, but they do not provide a mechanism to reconstruct the corrupted contents. 
Self-recovery image watermarking methods, on the other hand, attempt to encode sufficient visual information into the host image to enable reconstruction of tampered areas~\cite{bouarroudj2025secure,shen2024hidingface,bouarroudj2025medical}.  
However, these methods often require encoding a substantial amount of information, 
resulting in significant watermark capacity request and quality degradation of watermarked contents.

To overcome these limitations, we propose a \textbf{multi-scale latent quantization watermarking framework} that embeds multi-scale, quantized latent information within the host image.  
Our method preserves the semantics of protected images while maintaining robustness to common degradations, without requiring any prior knowledge of the tampering process.

Importantly, our framework is designed to be \textbf{plug-and-play compatible} with existing watermarking paradigms: it can be seamlessly integrated into both \emph{post-hoc watermarking} pipelines (e.g., EditGuard \cite{zhang2024editguard}) and \emph{in-generation watermarking} systems (e.g., Gaussian Shading \cite{yang2024gaussianshading}), allowing flexible deployment across different protection scenarios.

In addition, we introduce \textbf{ImageNet-S}, a benchmark dataset that provides paired image–label retrieval tasks for evaluating both tampered and recovered images. 
This benchmark facilitates quantitative analysis of recovery accuracy and semantic consistency, offering a standardized platform for future research in tampering recovery and watermarking-based restoration.

In summary, our contributions are threefold:
\begin{itemize}
    \item We introduce a \textbf{multi-scale latent watermarking strategy} that achieves robust, low-overhead embedding for image self-recovery.
    \item We design a \textbf{quantization-based hiding mechanism} that balances capacity, imperceptibility, and robustness against compression and noise.
    \item We demonstrate that our framework is \textbf{plug-and-play compatible} with both post-hoc and in-generation watermarking schemes, allowing flexible integration across protection pipelines.
    \item We establish \textbf{ImageNet-S}, a benchmark that enables consistent evaluation of image recovery and label retrieval for tampered and restored images.
\end{itemize}


\section{Related Work}
\label{sec:related_work}

In this paper, we introduce prior studies on image deepfake detection, localization, and recovery.

\subsection{Deepfake Detection and Localization}  
Beyond ownership verification, deepfake detection and tampering localization have also attracted significant attention.  
FakeShield~\cite{xufakeshield} leverages large language models to both identify manipulated regions and provide textual explanations for tampering.  
EditGuard~\cite{zhang2024editguard} jointly embeds watermark and image content within a unified representation, enabling both ownership verification and fine-grained tampering localization.  
Similarly, VideoShield~\cite{huvideoshield} extends Gaussian Shading~\cite{yang2024gaussianshading} to the video domain, using latent noise consistency to detect and localize edits across frames.

However, most existing methods focus primarily on ownership protection or localization, leaving the recovery of the original, untampered content largely unexplored.  

\subsection{Deepfake Recovery}  
Without using watermarking, DFREC~\cite{yu2024DFREC} learns to directly reconstruct the original image from a tampered input using supervised training on paired data (original–tampered image pairs).  
However, it requires explicit knowledge of tampering regions and manipulation types, limiting its practicality in real-world scenarios.  
HidingFace~\cite{shen2024hidingface} attempts to preserve facial integrity by hiding facial information in background regions and recovering it during restoration.  
Similar to DFREC, it relies on known tampering locations and embeds a large amount of information (proportional to the face region size), making it susceptible to information loss under common image degradations.


\subsection{Self-Recovery Image Watermarking}
Self-recovery (or self-embedding) watermarking \cite{amrullah2026tamperSurvey,sisaudia2024approximate,bouarroudj2025medical} methods aim to both detect tampering and restore tampered regions by embedding compact digests of image blocks back into the image. Typically, an image is divided into non-overlapping blocks, and each block generates two types of information:
1) Authentication bits, used for tamper detection.
2) Recovery bits, representing a compressed version of the original block (e.g., most significant bits (MSBs), block averages, or transform-domain coefficients).
These bits are embedded into other blocks according to a pre-defined mapping strategy, enabling reconstruction when tampering is detected.

Despite their utility, self-recovery methods generally rely on \emph{fragile} watermarking mechanisms, such as replacing least significant bits (LSBs) with watermark bits, due to the need to embed large amounts of image information. 
This dense embedding (typically 2–4 bits per pixel (bpp)) makes them highly sensitive to even mild lossy compression \cite{wan2022comprehensive}, which can corrupt the embedded information and compromise recovery.
The total embedded bit count can be expressed as 
$\text{bpp} \times H \times W \times 3$ for a color image of size $H\times W$.  
For instance, a $256{\times}256{\times}3$ image with $\text{bpp}=3$ requires $589{,}824$ embedded bits—an impractically large payload that undermines robustness in real world.

Moreover, the \emph{coincidence problem}~\cite{tai2018coincidenceProblem} further limits recovery reliability: since each block’s recovery data are typically stored in another block, simultaneous tampering of both regions leads to unrecoverable loss.  
As a result, conventional self-recovery watermarking remains fragile and poorly suited to realistic conditions involving complex tampering, lossy transmission, or compression artifacts.
\section{Preliminary}
\label{sec:preliminary}

In the image domain, Vector Quantized Variational Autoencoders (VQ-VAEs) \cite{van2017vqvae,esser2021VQGAN,tian2024VAR} represent images as discrete latent tokens rather than continuous pixel values. 
The encoder maps an image into a grid of quantized indices drawn from a finite codebook, and these tokens can then be modeled autoregressively to enable image synthesis. 
Although two main autoregressive strategies, next-token prediction \cite{van2017vqvae,esser2021VQGAN} and next-scale prediction \cite{tian2024VAR}, are commonly employed for generating such latent tokens, 
they differ in the spatial context during generation.
Please see Sec. \ref{Sec: Latent Modeling} of Appendix for details.

\section{Proposed Method}\label{sec:method}

In this section, problem formulation is first described in Sec. \ref{sec:problem_formulation}. 
To enhance the robustness of hidden information, we represent each image as discrete multi-scale token maps, as described in \cref{sec:quantization_watermarking}, and demonstrated the plug-and-play compatibility of our method with various watermarking and localization frameworks in \cref{sec:watermark_localization}.
In Sec. \ref{Sec: Self-Recovery}, recovery of tampered images is described.

\subsection{Problem Formulation}
\label{sec:problem_formulation}
Given a tampered image $I_d$ that has been manipulated from a clean image $I$, our objective is to generate image $I_r$ that is recovered from $I_d$ to closely approximate the original $I$, and subsequently retrieve from a target dataset $\mathcal{D} = \{(x_i,y_j), i=1,\ldots,N\}$ with $I_r$ as the input, where $N$ is the number of images in the dataset and $(x_i,y_j)$ denotes a pair of image and its corresponding label.


\subsubsection{Proactive Recovery}\label{Sec: Proactive-Recovery}
The proactive approach assumes that the image owner preprocesses the clean image $I$ into a protected version $I_w$ before distribution.
After potential tampering, the fake image $I_d$ modified from $I_w$ can then be recovered to $I_r$ using the self-embedded information within $I_w$.

\subsubsection{Limitations and Our Motivation}
\label{sec:motivation}
Most existing tampering recovery methods \cite{shen2024hidingface,yu2024DFREC} focus exclusively on facial regions, where the tampered area is relatively constrained and predictable.
Notably, \cite{yu2024DFREC} requires learning the differences between clean and tampered images using paired data ($I$, $I_d$), which is an impractical assumption in real-world scenarios.
In contrast, our goal is to recover general natural images, where the tampering regions and deepfake types are unknown and potentially arbitrary.
Moreover, conventional self-recovery approaches \cite{amrullah2026tamperSurvey,sisaudia2024approximate,bouarroudj2025medical,bouarroudj2025secure} often rely on embedding large amounts of information within the image itself, which significantly compromises robustness.
To address these challenges, we propose a robust and scalable recovery framework that employs watermark-based protection for the clean image, enables tampering localization to identify manipulated regions without prior knowledge of manipulated regions, and leverages multi-scale quantization to efficiently reduce the amount of information required for deepfake recovery.

\subsubsection{Factual Retrieval}
\label{sec:retrieval_accuracy}
To evaluate the recovery quality of reconstructed image $I_r$ and its retrievability from a target dataset $\mathcal{D} = \{(x_i,y_j), i=1,\ldots,N\}$, we measure the semantic similarity between $I_r$ and each image $x_i \in \mathcal{D}$.
Following the retrieval evaluation protocols in~\cite{liu2023E4S, baliah2025realistic}, we employ the CLIP cosine similarity~\cite{radford2021clip} as the retrieval metric to locate the original image $I$ based on its recovered counterpart $I_r$ by
\begin{equation}\label{eq:clip_similarity}
    s(x_i, I_r) =  \frac{M_\text{clip}(x_i) \cdot M_\text{clip}(I_r)} {\|M_\text{clip}(x_i)\|_2 \|M_\text{clip}(I_r)\|_2},
\end{equation}
where $s(x_i, I_r)$ denotes the similarity score between $x_i$ and $I_r$, and $M_\text{clip} : \mathcal{R}^{H \times W \times 3} \rightarrow \mathcal{R}^{d_e}$ is the CLIP image encoder that projects an image into a $d_e$-dimensional embedding space.

To quantify retrieval performance, we compute Top-k image accuracy, which measures whether the ground-truth clean image $I$ appears among the top-k most similar images in the dataset according to the similarity scores.
Formally, let Top-$k(I_r)$ denote the set of k images with the highest similarity scores to $I_r$:
\begin{equation}\label{eq:topk_image}
\text{Top-}k(I_r) =
\underset{x_i \in \mathcal{D}}{\text{arg top-}k}\ s(x_i, I_r).
\end{equation}
Then, the Top-k accuracy is defined as:
\begin{equation}\label{eq:topk_accuracy}
\text{Acc@}k =
\frac{1}{N} \sum_{n=1}^{N}
1 \left[ I_n \in \text{Top-}k(I_{r,n}) \right],
\end{equation}
where $1[\cdot]$ is the indicator function that equals $1$ if the ground-truth $I_n$ is successfully retrieved within the top-k results from its recovered counterpart $I_{r,n}$, and 0 otherwise.

This retrieval accuracy formulation parallels the inverse biometrics problem described in~\cite{gomez2020reversing}, where an adversary aims to reconstruct an unprotected template into a synthetic sample that matches a genuine one.
As noted in~\cite{gomez2020reversing}, the original image $I$ typically exists in the dataset during the development stage but may not always be present in the target dataset $D$ during validation or inference.
Therefore, we also consider a more general scenario, where $I \notin D$, in which case we evaluate Top-k label accuracy—measuring whether the retrieved images share the same semantic label as $I$ rather than matching its exact instance.

\subsection{Multi-Scale Self-Watermarking}
\label{sec:quantization_watermarking}
\begin{figure*}[t]
    \centering   
    \includegraphics[width=0.8\linewidth]{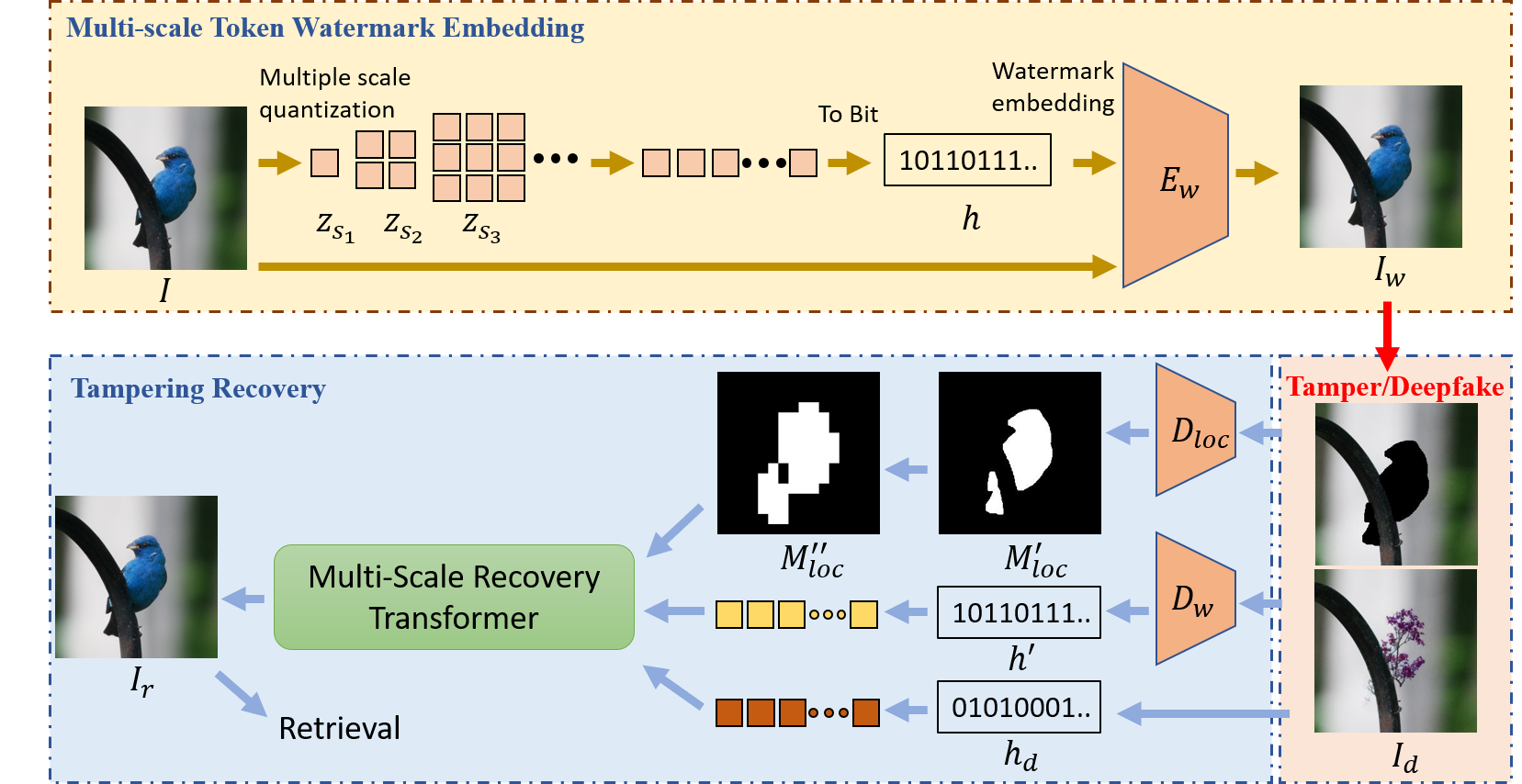}    
    \caption{Overall workflow of the proposed method. 
    In the multi-scale token watermarking stage, our goal is to protect an image $I$ by embedding image content–related information to produce a watermarked image $I_w$. The original image $I$ is first converted into quantized multi-scale tokens $z_{s_1}, \ldots, z_{s_k}$ using a VQ-VAE. These tokens are then flattened and transformed into a bit string $h$. The bit string $h$ is embedded into $I$ as a content watermark through the watermark injection encoder $E_w$, where the embedding length is constrained by the watermark capacity $|m|$ such that $|h| \leq |m|$. When a tampered image $I_d$ is generated from the watermarked image $I_w$ by malicious manipulations (e.g., object removal or inpainting), our method enables deepfake recovery using the content-related information embedded in $I_w$. The tampered image $I_d$ is decoded by the watermark decoder $D_w$ to extract the hidden quantized multi-scale tokens $h'$ and by the localization decoder $D_{loc}$ to produce a deepfake localization map $M_{loc}$. Finally, the multi-scale recovery transformer reconstructs the tampered image $I_d$ based on $h'$, $M_{loc}$, and the quantized tokens $h_d$ of the tampered image.}
    \label{fig:workflow}
\end{figure*}

To achieve proactive recovery described in Sec. \ref{Sec: Proactive-Recovery}, directly hiding the entire image $I$ itself is impractical due to its large data volume, which reduces embedding robustness and increases the risk of information loss during tampering.
To address this, we employ the multi-scale quantization mechanism \cite{tian2024VAR} described in \cref{sec:next_scale_prediction} to represent the image compactly as a set of token maps $(z_{s_1},\ldots,z_{s_K})$.
The corresponding workflow can be found in the upper part of \cref{fig:workflow}.
To embed these representations efficiently, we transform the quantized tokens into a binary bitstream based on their corresponding codebook indices.
This allows us to conceal the essential image semantics rather than raw data, significantly reducing the amount of embedded information while maintaining recoverability.
Formally, the hidden information $h$ is constructed as:
\begin{equation}\label{eq:hiding_information}
\begin{split}
h = \text{bin}\Big( 
    \text{concat}\big( 
        \text{flatten}\big(\text{index}(z_{s_1}, \mathcal{C})\big), \ldots, \\
        \text{flatten}\big(\text{index}(z_{s_K}, \mathcal{C})\big), 
    \big) 
\Big)
\end{split}
\end{equation}
where $\text{index}(z_{s_i}, \mathcal{C})$ maps each quantized token in $z_{s_i}$ to the corresponding codebook index,
$\text{flatten}(\cdot)$ vectorizes the token grid,
$\text{concat}(\cdot)$ concatenates all scales, and
$\text{bin}(\cdot)$ converts the concatenated indices into a binary bitstring.
Then, the protected image $I_w$ is generated by embedding the bitstring $h$ into the original image using a watermarking encoder $E_w(\cdot)$:
\begin{equation}\label{eq:watermarked_image}
I_w = E_w(I, h).
\end{equation}

Since the embedding capacity of the watermarking model is limited to a predefined maximum bit length $|m|$, the total length of the hidden bitstream must satisfy:
\begin{equation}\label{eq:capacity_constraint}
|h| \leq |m|.
\end{equation}
To meet this constraint, we only use the first $k$ scales of the token hierarchy $h=(z_{s_1}, \ldots, z_{s_k})$, where $k < K$ is selected such that $|h|$ does not exceed the available embedding capacity.

\begin{figure*}[t]
    \centering
    \begin{subfigure}{\textwidth}
        \centering
        \includegraphics[width=\textwidth]{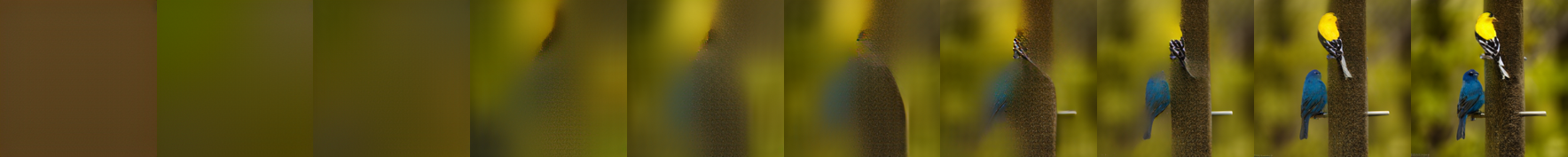}
        \caption{Reconstructed image via multi-scale quantization without dropout (VAR \cite{tian2024VAR}).}
        \label{fig:dropout_explain_wo_dropout}.
    \end{subfigure}       

    \begin{subfigure}{\textwidth}
        \centering
        \includegraphics[width=\textwidth]{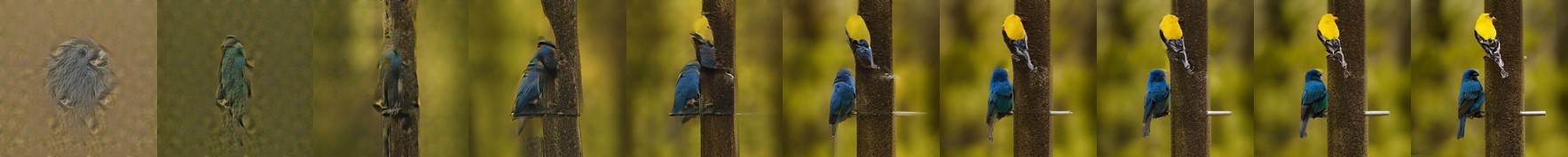}
        \caption{Reconstructed image via multi-scale quantization with dropout.}
        \label{fig:dropout_explain_w_dropout}.
    \end{subfigure}    
    \caption{Visualization of the dropout effect during quantizer (VQ-VAE) training. From left to right, each column shows the reconstructed images using token maps $(z_{s_1}, \ldots, z_{s_k})$ with $1\leq k\leq 10$.  
    Dropout encourages semantic robustness across scales, allowing lower-level representations (smaller $k$) to retain more global structure and meaningful content.  
    In contrast, the model trained without dropout produces coarse, less informative reconstructions at lower scales.}
    \label{fig:dropout_explain}
\end{figure*}

However, as shown in \cref{fig:dropout_explain_wo_dropout}, the quantizer from VAR~\cite{tian2024VAR} tends to concentrate most semantic information in the last few scales ({\em e.g.}, $z_{s_{K-1}}$ and $z_{s_K}$).  
While this is acceptable for standard image reconstruction or generation, it is unsuitable for our purpose, as we require meaningful representations even at lower scales.  
To distribute semantic information more evenly across scales, we adopt the dropout-based training strategy \cite{kumar2023rvqgan}.  
During VQ-VAE training, the last several scales are randomly dropped with a dropout rate $0.1$, forcing the quantizer to encode informative features at smaller scales as: $(z_{s_1}, \ldots, z_{s_n})$ for $1 \leq n \leq K$.
As shown in \cref{fig:dropout_explain_w_dropout}, the semantics appear in the early scales compared to original VAR without dropout (\cref{fig:dropout_explain_wo_dropout}).

The capacity constraint in \cref{eq:capacity_constraint} also prevents the use of single-scale quantization schemes, such as those used in next-token prediction frameworks~\cite{esser2021VQGAN, van2017vqvae}, since they fail to effectively compress the entire image representation.  
On the other hand, the limitations of single-scale quantization for recovery watermarking are discussed in Sec. \ref{Sec: Latent Modeling} of Appendix to motivate our design of multi-scale self-watermarking here.

Overall, these limitations explain our motivation of studying \emph{multi-scale quantization} here, which strikes a balance between compactness and fidelity by encoding essential semantic content across scales while satisfying the watermarking capacity constraint.


\subsection{Plug-and-Play Characteristic}
\label{sec:watermark_localization}
In this section, our method is demonstrated to enable seeming integration in a \emph{plug-and-play} manner with a wide range of watermarking-based deepfake detection frameworks, including both \textit{in-generation} ({\em e.g.},~\cite{wen2023treering, ci2024ringid, yang2024gaussianshading, mao2025maxsive, fang2025syntag}) and \textit{post-hoc} ({\em e.g.},~\cite{zhu2018hidden, zhang2024editguard, zhang2025omniguard}) watermarking approaches.

\paragraph{Integration with Post-Hoc Watermarking.}
For demonstration purposes, we adopt EditGuard~\cite{zhang2024editguard} as the baseline, as it provides an integrated deepfake localization module alongside watermarking.
Importantly, EditGuard can be replaced with more advanced watermarking framework~\cite{zhang2025omniguard}, and the localization component operates independently of the watermarking process.
Thus, employing a state-of-the-art localization model~\cite{xufakeshield, huang2025sida} can further improve recovery performance without modifying the core of our framework.

Our pipeline first embeds the localization watermark $W_{\text{loc}}$ into the original image, and then hides the binary bitstream $h$ (\cref{eq:watermarked_image}) for tampered image recovery.  
The resulting protected image $I_w$ is obtained as:
\begin{equation}\label{eq:container_image}
I_w = E_w(E_{\text{loc}}(I, W_{\text{loc}}),\, h),
\end{equation}
where $E_{\text{loc}}$ denotes the localization embedding network and $W_{\text{loc}}$ is initialized as a purely blue image for easily determining the tampering mask $M_{\text{loc}}'$.

If the watermarked image $I_w$ is tampered with to produce a manipulated version $I_d$, both the hidden image representation and tampering localization map can be extracted as:
\begin{equation}\label{eq:decoding_process}
M_{\text{loc}}' = D_{\text{loc}}(I_d), \qquad
h' = D_w(I_d),
\end{equation}
where $D_{\text{loc}}$ is the localization decoder that predicts the tampering mask $M_{\text{loc}}'$, and $D_w$ is the watermark extractor that retrieves the hidden bitstream $h'$.

\paragraph{Integration with In-Generation Watermarking.}

Gaussian Shading-based approaches~\cite{yang2024gaussianshading, mao2025maxsive, fang2025syntag} employ Stable Diffusion~\cite{rombach2022StableDiffusion} as the generative backbone, where the watermark is embedded by adjusting the initial noise $x_T$ according to a given bit string before image synthesis.  
This setting differs from ours, as the hidden information must be determined \emph{prior} to image generation.  
Inspired by~\cite{muller2025black}, we address this by introducing an optimization process that adapts our hiding mechanism to this scenario.

Specifically, Stable Diffusion generates an image $I_g$ and its latent $x_0$ from a random initial noise $x_T$.  
In our setting, we discard the original noise $x_T$ and instead aim to estimate a new latent noise $x_T'$ that encodes the desired hidden information $h$, obtained through the quantization of $I_g$.
For this, we perform an optimization based on the DDIM inversion process, aligning the forward diffusion trajectory with the target latent $x_T'$.
Formally, the optimization objective is defined as:
\begin{equation}\label{eq:optimization}
\mathcal{L}_{\text{mse}} = \| DDIM_{0 \rightarrow T}(x_0 + \delta) - x_T' \|_2^2,
\end{equation}
where $DDIM_{0 \rightarrow T}$ denotes the deterministic inversion from timestep $0$ to $T$, and $\delta$ is a learnable perturbation optimized to ensure the latent $x_0$ produces a diffusion trajectory consistent with the target noise $x_T'$.

This optimization enables our method to operate in conjunction with in-generation watermarking pipelines, effectively bridging pre-generation watermark embedding and post-generation recovery.
Overall, the modular design of our framework ensures compatibility with both existing and future watermarking and localization techniques without requiring architectural modifications.


\subsection{Tampering Recovery}\label{Sec: Self-Recovery}
As illustrated in the bottom part of \cref{fig:workflow}, we describe the recovery process, which reconstructs the original image from the tampered input $I_d$ using the extracted hidden information $h'$ and the predicted tampering localization mask $M_{\text{loc}}'$ obtained from \cref{eq:decoding_process}.

\subsubsection{Patchification of the Localization Mask}
As discussed in \cref{sec:next_scale_prediction,eq:hiding_information}, the hidden information $h$ represents a compressed form of the original image $I \in \mathcal{R}^{H \times W \times 3}$, where pixel-level information is first encoded into a feature map $f \in \mathcal{R}^{h \times w \times C}$ and subsequently quantized into discrete token indices $(z_{s_1}, \ldots, z_{s_K})$.
The quantization operates on image patches, where each token corresponds to a spatial region of size $p=\frac{H}{h}=\frac{W}{w}$. 

Therefore, to align the predicted localization mask $M_{\text{loc}}' \in [0,1]^{H \times W}$ with the token-level representation, we downsample it into a patch-level mask $M_{\text{loc}}'' \in \{0,1\}^{h \times w}$ using average pooling followed by a thresholding operation:
\begin{equation}\label{eq:localization_patchify}
\tilde{M}_{\text{loc}}''(i, j) =
\frac{1}{p^2} \sum_{u = i \cdot p}^{(i+1) \cdot p - 1}
\sum_{v = j \cdot p}^{(j+1) \cdot p - 1}
M_{\text{loc}}'(u, v),
\end{equation}
\begin{equation}\label{eq:localization_threshold}
M_{\text{loc}}''(i, j) =
\begin{cases}
1, & \text{if } \tilde{M}_{\text{loc}}''(i, j) \ge \tau, \\
0, & \text{otherwise},
\end{cases}
\end{equation}
where $\tau \in [0,1]$ is a predefined threshold controlling the sensitivity of tampering detection at the patch level.

This thresholded, patchified mask $M_{\text{loc}}''$ aligns the localization signal with the quantized token map, effectively bridging pixel-level tampering detection and token-level semantic representation.
By filtering out minor or uncertain tampering regions, it enables a more stable and interpretable recovery process in the subsequent reconstruction stage.

\subsubsection{Recovery with Conditional Transformer}
Due to the capacity limitation $|h| \leq |m|$ of the watermarking mechanism described in \cref{eq:capacity_constraint}, 
the embedded information $h = (z_{s_1}, \ldots, z_{s_k})$ only encodes a partial representation of the original image $I$, 
where $k < K$. 
To compensate for the missing semantic and structural details, 
we employ a Transformer decoder with the same architecture as VAR~\cite{tian2024VAR}, 
which enables hierarchical autoregressive reconstruction in the latent token space.

\begin{figure}[t]
    \centering   
    \includegraphics[width=\linewidth]{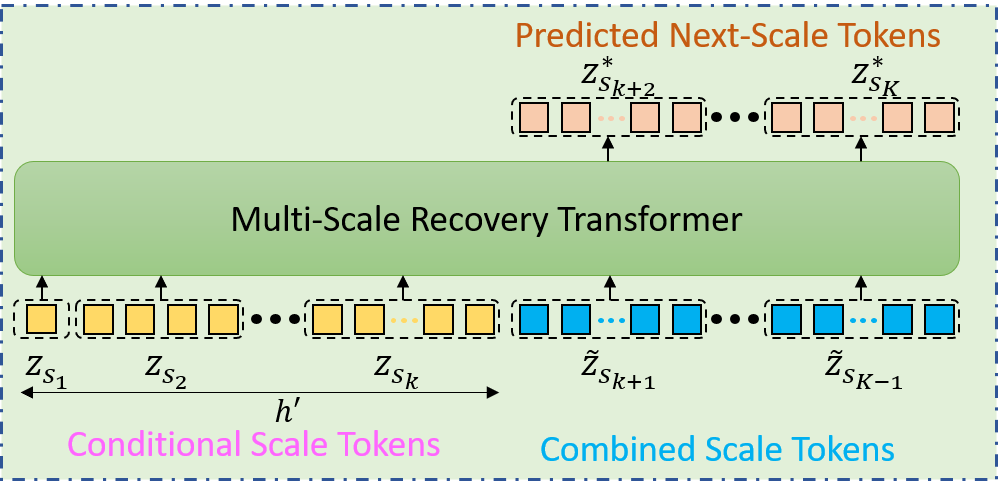}    
    \caption{
    Illustration of proposed conditional deepfake recovery Transformer. 
    The predicted hidden codes $h'$ are used as conditional tokens to guide the generation process, enabling context-aware reconstruction. 
    The untampered regions $\tilde{z}_{s_i}$ of deepfaked image provide additional contextual cues, helping the model produce more detailed and consistent predictions at higher scales.
    }
    \label{fig:conditional_transformer}
\end{figure}

We integrate the next-scale prediction, the conditional Transformer, 
and the patch-level localization map $M_{\text{loc}}''$ 
to achieve the recovery objective illustrated in \cref{fig:conditional_transformer}. 
Specifically, we extend the next-scale prediction in \cref{eq:next_scale_prediction} to a 
\emph{conditional next-scale prediction} that incorporates spatial localization cues 
from detected tampered regions.

\paragraph{Conditional Sequential Prediction and Fusion.}
Assume we have a verified clean prefix
$h = (z_{s_1}, \ldots, z_{s_k})$ 
and the full tampered representation 
$h_d = (z^d_{s_1}, \ldots, z^d_{s_K})$. 
Our objective is to reconstruct the missing or corrupted higher-scale token maps 
$(z_{s_{k+1}}, \ldots, z_{s_K})$ 
by sequentially predicting and fusing clean estimates with the tampered tokens 
under the guidance of the patch-level localization mask 
$M_{\text{loc}}'' \in [0,1]^{h \times w}$.

For each subsequent scale $i = k{+}1, \ldots, K$, 
we first predict the clean token map conditioned on all previously known or recovered scales:
\begin{equation}\label{eq:conditional_next_scale_pred}
z_{s_i}^{*} = f_{\theta}\big(z_{s_1}, \ldots, z_{s_k}, \tilde{z}_{s_{k+1}}, \ldots, \tilde{z}_{s_{i-1}}\big),
\end{equation}
where $f_{\theta}$ denotes the conditional next-scale predictor implemented by the Transformer decoder.

Next, we fuse the predicted token $z_{s_i}^{*}$ with the corresponding tampered token $z^d_{s_i}$ 
through a convex combination guided by the localization mask:
\begin{equation}\label{eq:conditional_next_scale_fusion}
\tilde{z}_{s_i} = (1 - M_{\text{loc}}'') \odot z^d_{s_i} + M_{\text{loc}}'' \odot z_{s_i}^{*},
\end{equation}
where $\odot$ denotes element-wise multiplication with spatial broadcasting of $M_{\text{loc}}''$ 
to match token dimensions. 
This fusion preserves intact regions from the tampered input 
while selectively restoring corrupted areas with model-predicted clean features.

By iteratively applying 
\cref{eq:conditional_next_scale_pred,eq:conditional_next_scale_fusion} 
for $i = k{+}1, \ldots, K$, 
we obtain the recovered latent representation:
\begin{equation}\label{eq:recovered_latent}
h^{\text{rec}} = (z_{s_1}, \ldots, z_{s_k}, z_{s_{k+1}}^{*}, \ldots, z_{s_K}^{*}).
\end{equation}
Finally, the recovered image is reconstructed by decoding the hierarchical token representation:
\begin{equation}\label{eq:recovered_image}
I_r = D(h^{\text{rec}}),
\end{equation}
where $D(\cdot)$ denotes the decoder of multi-scale quantization.

\section{Experimental Results}
\label{sec:experiments_and_results}
Main results are presented below and ablation studies are shown in Sec. \ref{Sec: More Results} of Appendix.

\begin{table*}[h]
    \caption{Comparison of retrieval accuracy and CLIP score between our method and related approaches. 
        Deepfake images are generated from the ImageNet-S dataset using Stable Diffusion. 
        “Same samples” indicates that the original image is included in the evaluation dataset, whereas “Different samples” indicates that the original image is excluded, allowing retrieval only within the same class. 
        “Top-$k$ label” denotes the retrieval accuracy based on whether any of the top-$k$ retrieved images from ImageNet-S share the same label as the original image. 
        “Top-$k$ image” denotes the retrieval accuracy based on whether the original image itself appears among the top-$k$ retrieved images.
        Higher CLIP scores indicate better semantic consistency between the recovered and original images.}
    \centering\setlength{\tabcolsep}{4.5pt}\scalebox{0.70}{
        \begin{tabular}{l cccc | cc | c}
            \toprule
            ImageNet &\multicolumn{4}{c}{Same samples} & \multicolumn{2}{c}{Different samples}           & CLIP score ($\uparrow$)\\            
            Methods {\textbackslash} Metric & Top-1 label ($\uparrow$) & Top-5 label ($\uparrow$) & Top-1 image ($\uparrow$) & Top-5 image ($\uparrow$) & Top-1 label ($\uparrow$) & Top-5 label ($\uparrow$) &  \\
            \midrule
            HiNet \cite{jing2021hinet}          & 0.2393 & 0.3511 & 0.2017 & 0.2739 & 0.0742 & 0.0716 & 0.7255\\
            RePaint \cite{lugmayr2022repaint}   & 0.2821 & 0.3868 & 0.2571 & 0.3373 & 0.0717 & 0.1693 & 0.7854\\
            
            \midrule    
            VQGAN \cite{esser2021VQGAN} (codebook size $V$ = 1024)         & 0.3633 & 0.5516 & 0.2756 & 0.4243 & 0.1787 & 0.3699 & 0.7887\\
            VQGAN \cite{esser2021VQGAN} (codebook size $V$ = 16384)         & 0.5205 & 0.6895 & 0.4246 & 0.5925 & 0.2467 & 0.4535 & 0.8157\\
            VAR \cite{tian2024VAR}              & 0.6754 & 0.8125 & 0.5761 & 0.7400 & 0.4222 & 0.6547 & 0.8817\\
            \midrule  
            \textbf{hidden code (Ours)}  & \textbf{0.8480} & \textbf{0.9485} & \textbf{0.7680} & \textbf{0.9031} 
                & \textbf{0.4638} & \textbf{0.7266} & \textbf{0.9000}\\
            \textbf{hidden code + Transformer (Ours)}  & \textbf{0.8782} & \textbf{0.9612} & \textbf{0.8040} & \textbf{0.9206} 
                & \textbf{0.4966} & \textbf{0.7474} & \textbf{0.9032}\\
            \textbf{hidden code + conditional Transformer  (Ours)}  & \textbf{0.9231} & \textbf{0.9814} & \textbf{0.8744} & \textbf{0.9549} 
                & \textbf{0.4968} & \textbf{0.7467} & \textbf{0.9168}\\
            \bottomrule
        \end{tabular}}
    
    \label{tab:exp_retrieval}
\end{table*}

\begin{figure*}[t]
    \centering   
    \includegraphics[width=0.85\textwidth]{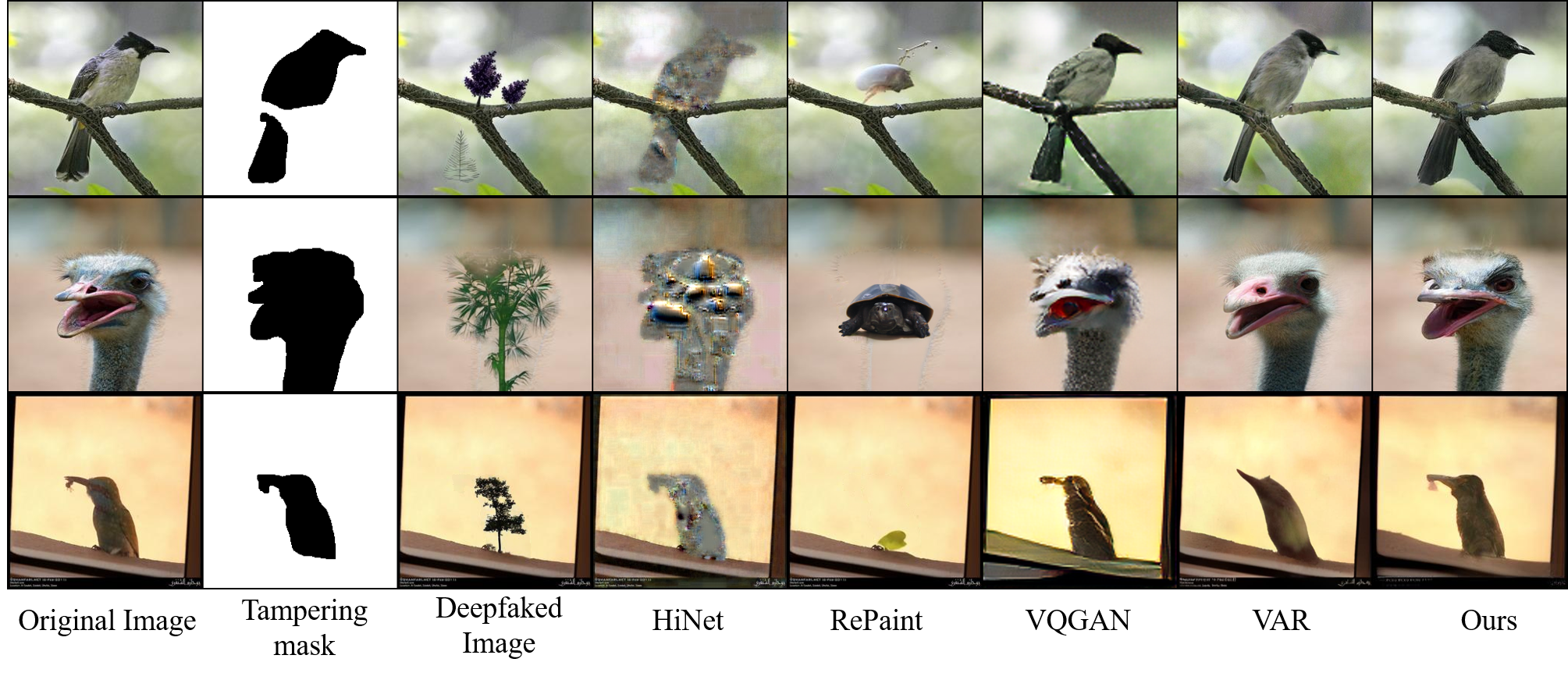} 
    \caption{Qualitative comparison of tampered image recovery performance. 
    The first two columns show the Original Image and its Tampering mask. The third column, Deepfaked Image, shows the tampered image generated using Stable Diffusion based on the mask.
    While general recovery methods (HINet \cite{jing2021hinet}, RePaint \cite{lugmayr2022repaint}) often fail to maintain semantic consistency or introduce severe distortions. 
    The general quantization methods (VQGAN, VAR) with our watermarking scheme struggle with texture preservation, our method successfully reconstructs the masked region, restoring the identity and realistic appearance of the original subjects (birds and an ostrich).    
    }
    \label{fig:all_recovery_visualization}
\end{figure*}

\begin{table*}[h]
    \caption{
        Comparison of different types of image recovery and watermarking approaches. 
        Traditional self-recovery watermarking methods embed large amounts of pixel or transform information into fragile regions such as LSBs, leading to high BPP and poor robustness. 
        In contrast, our method encodes compact hidden-code representations through robust, multi-scale quantization in the latent space, achieving efficient and resilient recovery.    
    }
    \centering\setlength{\tabcolsep}{4.5pt}\scalebox{0.70}{
        \begin{tabular}{lc ccc cc}            
            \textbf{Method} & \textbf{Type} & \textbf{Hiding Information} & \textbf{Smallest Hiding Unit} & \textbf{Hiding Domain} & \textbf{Bits per Pixel (BPP)} & \textbf{Total Hidden Information} \\
            \toprule
            HiNet~\cite{jing2021hinet} & Reversible module & Pixel-level content & $1\times1$ & Latent space & -- & $H \times W \times C$ \\
            RePaint~\cite{lugmayr2022repaint} & Inpainting model & -- & -- & -- & -- & -- \\
            \midrule 
            \makecell[l]{Common\\self-recovery watermarking ~\cite{amrullah2026tamperSurvey}} & Fragile watermark 
            &  \makecell[c]{MSBs, block averages,\\or transform-domain\\coefficients} 
            & $2\times2$--$6\times6$ & Pixel (LSB) & 2--4 & $\text{bpp} \times H \times W \times C$ \\
            Bouarroudj et al.~\cite{bouarroudj2025secure} & Fragile watermark & DWT coefficients & $4\times4$ & Frequency domain & 0.5 & $\text{bpp} \times H \times W \times C$ \\
            \midrule  
            \textbf{Ours} & Robust watermark & Multi-scale quantized vectors & $16\times16$ & Latent space & -- & $|h|$ ($<|m|$) \\
            
            \bottomrule
        \end{tabular}}
    
    \label{tab:comparison}
\end{table*}

\subsection{Dataset}
\label{sec:dataset}
To construct the experimental setup for the watermarking and recovery scenario described in \cref{sec:problem_formulation}, we require a dataset\footnote{No such dataset is available in the literature.} that provides triplets of the form 
$\{ (x_i, y_i, M_{\text{loc},i}) \}$,  
where $x_i$ denotes a clean image, $y_i$ is its semantic label, and $M_{\text{loc},i} \in \{0,1\}^{H \times W}$ represents the corresponding tampering localization (segmentation) mask.
We adopt the ImageNet dataset as the base source of clean images and labels ($\{x_i, y_i\}$), and generate the tampering localization masks $M_{\text{loc},i}$ using the LISA framework~\cite{lai2024lisa}, which provides high-quality, object-level segmentation annotations. 
The resulting dataset, composed of triplets 
$\{ (x_i, y_i, M_{\text{loc},i}) \}$, 
is referred to as \textbf{ImageNet-Segment (ImageNet-S)}.
This dataset serves as a benchmark for assessing both the effectiveness of information hiding and retrieval performance under controlled tampering conditions.

\subsection{Experimental Setup and Implementation}
We selected two representative watermarking and localization approaches as our base models to cover both \textit{post-hoc} and \textit{in-generation} watermarking paradigms.  
For the post-hoc watermarking approach, we trained EditGuard with a watermark capacity of $|m| = 1024$, which jointly functions as a watermark encoder and tampering localization model.  
For the in-generation watermarking baseline, we adopted the Gaussian shading-based VideoShield framework~\cite{huvideoshield}, using parameters $k_c = 1$, $k_h = k_w = 4$, and $k_f = 1$, with a corresponding watermark length of $|m| = 1024$, to embed watermarks directly during image generation.

In our implementation, only the first $|h|$ bits of the watermark were used to conceal image-related information, while the remaining bits were set to zero. 
We chose a relatively high watermark capacity ($|m| = 1024$) to ensure that the same watermarking model can accommodate different lengths ($|h|$) of hidden information and optionally reserve capacity for ownership or provenance watermarking when needed.  
If desired, this capacity can be reduced to $|m| = |h|$ without loss of generality.

To prevent tampered regions from unduly influencing the recovered image $I_r$, we apply a conservative threshold $\tau = 0.05$ to the soft localization map $\tilde{M}_{\text{loc}}''(i, j)$.  
All tampering operations were performed using the diffusion-based image editing framework Stable Diffusion~\cite{rombach2022StableDiffusion}.
We note that the official implementations of DFREC~\cite{yu2024DFREC} and HidingFace~\cite{shen2024hidingface} are not publicly available.  
Since HidingFace is based on the HiNet architecture~\cite{jing2021hinet}, we include HiNet as a representative baseline for comparison.

\subsection{Factual Retrieval}
\label{exp:retrieval_imagenet}
We evaluated the proposed framework on ImageNet-S to assess factual retrieval accuracy, comparing it against both inpainting-based recovery (e.g., RePaint \cite{lugmayr2022repaint}) and information-hiding method (e.g., HiNet \cite{jing2021hinet}).
As shown in Table \ref{tab:exp_retrieval}, our method consistently outperforms these baselines across all retrieval metrics, highlighting the effectiveness of our hidden-code representation.

We further evaluated our method under different quantization backbones, including single-scale quantization (VQGAN) and multi-scale quantization (VAR).  
Our model variants demonstrate progressive improvements as we introduce more structured recovery mechanisms:  
(1) the baseline \textbf{hidden code} achieves strong retrieval accuracy,  
(2) the addition of a \textbf{Transformer} enhances hierarchical feature reconstruction, and  
(3) the \textbf{conditional Transformer}, guided by the localization map, achieves the best overall performance with a Top-1 label accuracy of 0.9231 and a Top-1 image retrieval accuracy of 0.8744.
These results confirm that our conditional recovery strategy effectively restores tampered regions and preserves global semantic integrity, as further evidenced by the high CLIP similarity scores.

\subsection{Visualization of Deepfake Recovery}
We show the qualitative comparisons corresponding to different methods mentioned in \cref{exp:retrieval_imagenet} in \cref{fig:all_recovery_visualization}, which shows the original image, its binary mask, and the reconstructed results produced by competing models and our approach.
While prior methods (e.g., HiNet \cite{jing2021hinet}, RePaint \cite{lugmayr2022repaint}) tend to generate texture artifacts or semantic drift, our model produces visually coherent reconstructions that faithfully recover both structural and contextual details.
We also show that our recovery scheme can be applied to different quantization methods, including single-scale quantization (VQGAN \cite{esser2021VQGAN}) and multi-scale quantization (VAR \cite{tian2024VAR}), by changing the backbone quantization model.
Notably, our approach restores fine object semantics and confirm the effectiveness of multi-scale latent modeling.

\begin{table}[h]
    \caption{
    Comparison of bit accuracy between the self-recovery watermarking method~\cite{bouarroudj2025secure} 
    and our proposed approaches (Ours(P) denotes the use of post-hoc watermarking model--EditGuard and Ours(I) denotes the use of in-generation model--Gaussian shading) under various image degradations. 
    Higher values indicate better robustness in bit preservation.
    }
    \centering\setlength{\tabcolsep}{4.5pt}\scalebox{0.70}{
        \begin{tabular}{l ccc cc}
            \toprule
            Methods & Clean ($\uparrow$) &  JPEG($\uparrow$) & Gaussian Blur($\uparrow$) & \multicolumn{2}{c}{Gaussian Noise }($\uparrow$)\\
            && QF $70\%$ & $5\times 5$ & $\sigma=1$ & $\sigma=5$\\

            \midrule
            Bouarroudj \textit{et al.}~\cite{bouarroudj2025secure} & 1.0000 & 0.5054 & 0.5042 & 0.5851 & 0.5047\\   
            \midrule  
            \textbf{Ours(P)} & 0.9988 & 0.9923 & 0.9983 & 0.9983 & 0.9577 \\
            \textbf{Ours(I)} & 0.9999 & 0.9951 & 0.9971 & 0.9971 & 0.9951 \\
            \bottomrule
        \end{tabular}}
    
    \label{tab:exp_robustness_metric}
\end{table}

\subsection{Comparison with Self-Recovery Approaches}
We evaluate the robustness of our method against traditional self-recovery watermarking~\cite{bouarroudj2025secure} and provide comparisons with other recovery methods discussed in \cref{exp:retrieval_imagenet}.

\textbf{Efficiency of embedding.}
\Cref{tab:comparison} summarizes the key differences between traditional self-recovery methods and our approach.
Conventional methods embed a large volume of pixel- or transform-level information, resulting in high bits-per-pixel (BPP) and substantial total hidden information ($\text{BPP} \times H \times W \times C$), which makes them highly sensitive to degradation.
In contrast, our method encodes compact multi-scale quantized latent vectors into the host image, reducing the total information volume while maintaining robust and accurate recovery.

\textbf{Robustness against attacks.}
Following \cite{zhang2024editguard}, we evaluated the Bit Accuracy (BA) of recovered watermarks across a set of common image corruptions in \Cref{tab:exp_robustness_metric}.
Traditional self-recovery watermarking suffers significant performance degradation, whereas 
both variants of our method — using post-hoc watermarking (EditGuard) or in-generation watermarking (Gaussian Shading) — maintain near-perfect bit accuracy across all tested degradations.

\section{Conclusion}
\label{sec:conclusion}
In this paper, we propose a deepfake image detection, localization, and recovery mechanism with focus on recovery of tampered image for factual retrieval.
Our method is demonstrated to be plug-and-play compatible with both post-hoc and in-generation watermarking schemes, allowing flexible integration across protection pipelines.
We also establish ImageNet-S, a benchmark that enables consistent evaluation of image recovery and factual retrieval for tampered and restored images.

\clearpage
{
    \small
    \bibliographystyle{ieeenat_fullname}
    \bibliography{main}
}

\clearpage
\setcounter{page}{1}
\maketitlesupplementary

\section{Modeling Discrete Latent Codes}\label{Sec: Latent Modeling}
\begin{figure}[th]
    \centering
        \includegraphics[width=.5\textwidth]{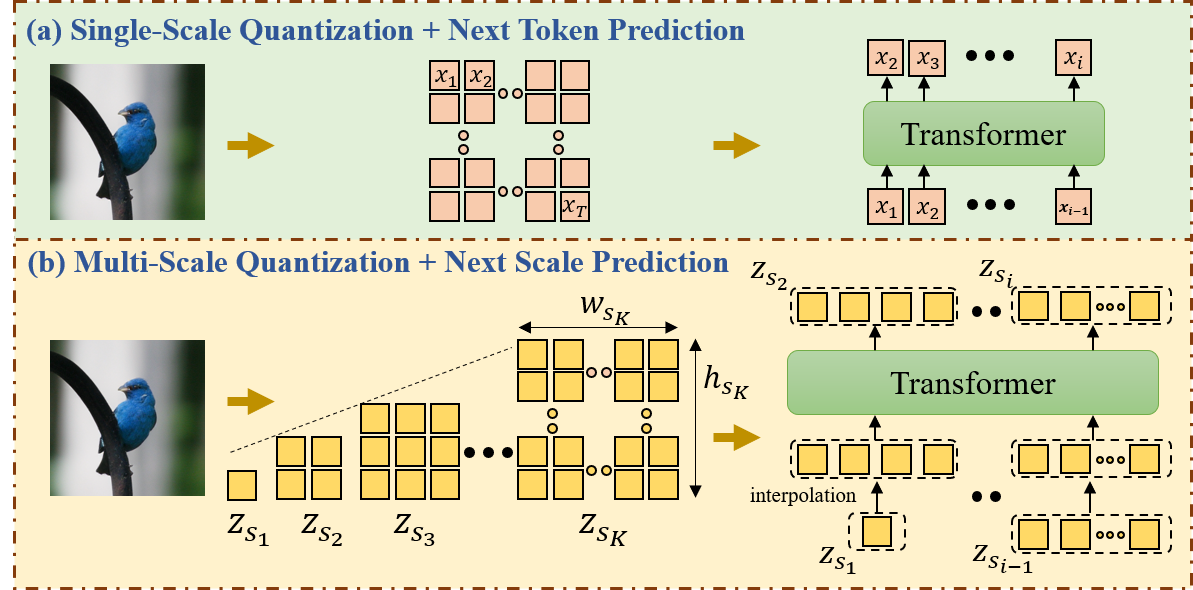}       
    \caption{    
    Illustration of single-scale versus multi-scale quantization.  
    (a) \textbf{Single-scale quantization:} each token $x_i$ corresponds to a specific spatial position in the image, and the transformer predicts $x_{i}$ sequentially based on all preceding tokens $(x_1,x_2, \ldots, x_{i-1})$.
    (b) \textbf{Multi-scale quantization:} each token map $z_{s_i}$ represents a group of tokens at a particular scale $s_i$.
    The transformer predicts $z_{s_i}$ conditioned on all previous token maps $(z_{s_1},z_{s_2}, \ldots, z_{s_{i-1}})$.    
    }
    \label{fig:quantization_explain}
\end{figure}

This section introduces the formulations of \emph{single-scale quantization} (Sec.~\ref{sec:next_token_prediction}) and \emph{multi-scale quantization} (Sec.~\ref{sec:next_scale_prediction}), whose differences are illustrated in Fig.~\ref{fig:quantization_explain}.  
We then describe how multi-scale quantization~\cite{tian2024VAR} enables content-dependent watermark embedding (Sec.~\ref{sec:quantization_watermarking}), and explain why single-scale quantization is unsuitable for tampering recovery (Sec.~\ref{Sec: Alternative-Recovery}).

\subsection{Next-Token Prediction} 
\label{sec:next_token_prediction}
Next-token prediction \cite{esser2021VQGAN,van2017vqvae} treats the quantized latent representation of an image $I$ as a one-dimensional sequence of discrete tokens $(x_1, x_2, \ldots, x_{T})$ (typically flattened from a two-dimensional latent grid), where $x_i$ represents the $i$-th token in the latent sequence, as illustrated in \cref{fig:quantization_explain} (a).
The model learns to predict each latent token sequentially, conditioned on all previously generated tokens:
\begin{equation}\label{eq:next_token_prediction}
    P(x_i | x_1,x_2, \ldots, x_{i-1}).
\end{equation}
This approach is analogous to pixel-by-pixel or word-by-word autoregressive modeling, as in PixelCNN \cite{van2017vqvae} and image-transformer architectures \cite{esser2021VQGAN}. The focus of next-token prediction is on local dependencies between neighboring tokens, enabling fine-grained reconstruction of spatial detail. However, since each token is predicted individually, this method can be computationally intensive and may struggle to capture large-scale spatial coherence across distant regions of an image.

\begin{figure*}[th]
    \centering
    \begin{subfigure}{.15\textwidth}
        \centering
        \includegraphics[width=\textwidth]{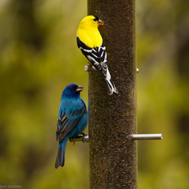}
        \caption{Original image.}
    \end{subfigure}       
    \begin{subfigure}{.6\textwidth}
        \centering
        \includegraphics[width=\textwidth]{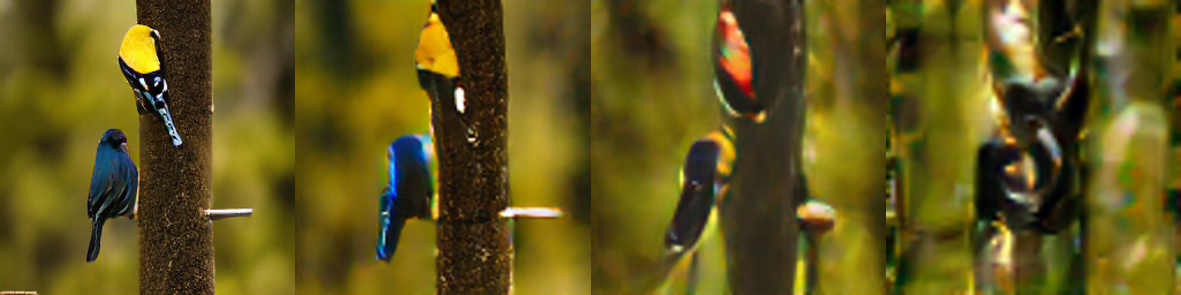}
        \caption{Reconstructed images via VQGAN at different resolutions.}
    \end{subfigure}    
    \caption{    
    Visualization of the limitations of single-scale tokenization at varying image resolutions.  
    (a) Original input image ($256{\times}256$).  
    (b) Reconstructed images using VQGAN, from left to right, correspond to image sizes of $256{\times}256$, $128{\times}128$, $96{\times}96$, and $64{\times}64$.  
    Even at $128{\times}128$, the reconstruction begins to lose fine details.  
    This degradation occurs because downsampling increases the information contained within each patch while the codebook remains fixed, forcing each quantized vector to represent a larger and more complex region.
    Consequently, the quantized representation becomes insufficient to capture fine-grained details, resulting in visible information loss and reduced visual fidelity.
    }
    \label{fig:vqgan_explain}
\end{figure*}

\subsection{Next-Scale Prediction}
\label{sec:next_scale_prediction}
Next-scale prediction \cite{tian2024VAR}, on the other hand, leverages the hierarchical structure of multi-scale VQ-VAE representations. 
This mechanism produces discrete latent grids at multiple spatial resolutions (e.g., coarse and fine scales), as illustrated in \cref{fig:quantization_explain} (b).

An input color image $I\in \mathcal{R}^{H \times W \times 3}$ is quantized into a hierarchy of $K$ multi-scale token maps, denoted as ($z_{s_1}$,$ z_{s_2}$,\ldots, $z_{s_K}$), where $z_{s_i}$ denotes the discrete token map at $s_i$ scale.
To construct these quantized representations, the image $I$ is first processed by VAE encoder to obtain a dense feature map $f_{s_1} \in \mathcal{R}^{h \times w \times C}$, 
where the spatial dimensions $h$ and $w$ are determined by the predefined patch size $p$ such that $h=\frac{H}{p}$ and $w=\frac{W}{p}$, and $C$ is the codebook embedding dimension.
This downscaling is inherently performed by the encoder’s convolutional hierarchy.

Initially, the feature map $f_{s_1}$ is progressively decomposed into a set of multi-scale latent representations by following a residual feature pyramid structure.
Given the discrete codebook $\mathcal{C} = \{ c_1, c_2, \ldots, c_V \} \subset \mathcal{R}^{V \times C}$, 
the feature map is first transformed into the lowest-scale feature representation $ r_{s_1} \in \mathcal{R}^{h_{s_1} \times w_{s_1} \times C}$.
Each subsequent scale $s_i$ increases spatial resolution, where $h_{s_i} \leq h_{s_{i+1}}$, $w_{s_i} \leq w_{s_{i+1}}$, and the finest scale satisfies $h_{s_K} = h$ and $w_{s_K} = w$.

The feature map $r_{s_i}$ is quantized to its nearest codebook entries, forming the discrete latent token map $z_{s_i}$.
The residual information is then computed by subtracting the reconstructed quantized features from the current feature map, yielding a refined feature $f_{s_{i+1}}$ for the next scale.
This process is iteratively repeated across scales, enabling hierarchical decomposition from coarse to fine representations.
Formally, the downsampling procedures can be written as:
\begin{equation}\label{eq:quantization1}
r_{s_i} = \text{downsample}(f_{s_i}, (h_{s_i}\times w_{s_i})),
\end{equation}
where $\text{downsample}(f_{s_i}, (h_{s_i}\times w_{s_i}))$ resizes the feature map $f_{s_i}$ to spatial resolution $\mathcal{R}^{h_{s_i}\times w_{s_i} \times C}$.
Each feature vector in $r_{s_i}$ is then quantized by mapping it to its nearest codebook entry:
\begin{equation}\label{eq:quantization}
z_{s_i}^{(p,q)} = \underset{c \in \mathcal{C}}{\arg\min} | r_{s_i}^{(p,q)} - c |_2,
\end{equation}
where $| \cdot |$ denotes the Euclidean distance and $(p,q)$ indexes spatial coordinates.

To maintain consistency across scales, the feature residual $f_{s_{i+1}}$ is updated by aggregating upsampled latent representations:
\begin{equation}\label{eq:quantization3}
f_{s_{i+1}} = f_{s_1} - \sum_{m=s_1}^{s_i} \text{upsample}(z_m, (h_{s_K}\times w_{s_K})),
\end{equation}
where $\text{upsample}(z_m, (h_{s_K}\times w_{s_K}))$ interpolates lower-resolution latent maps to match the highest scale $(h_{s_K}\times w_{s_K})$.
 
Since $z_{s_i}$ represents the quantized embedding of the feature map at scale ${s_i}$, capturing both semantic and spatial information with a resolution proportional to $h_{s_i}\times w_{s_i}$.
This multi-scale quantization enables hierarchical representation learning, allowing subsequent models to reason over coarse-to-fine image abstractions.

Building upon this structure, next-scale prediction models the conditional dependency between latent representations across scales as:
\begin{equation}\label{eq:next_scale_prediction}
    P(z_{s_i} \mid z_{s_1},z_{s_2}, \ldots, z_{s_{i-1}}),
\end{equation}
where $z_{s_{i-1}}$ denotes the coarser latent map corresponding to $z_{s_i}$ that is the finer one.
This formulation allows the model to first generate a rough global structure at a lower resolution and then refine it progressively by predicting finer details at higher resolutions. 
As a result, next-scale prediction captures global spatial dependencies and structural consistency more effectively than token-level autoregression.

\subsection{Single-scale quantization for Recovery Watermark Embedding}
\label{Sec: Alternative-Recovery}
In \cref{sec:quantization_watermarking}, we adopt multi-scale quantization \cite{tian2024VAR} as the backbone for embedding content-dependent self-watermarks.  
Here, we explain why classical single-scale quantization~\cite{van2017vqvae,esser2021VQGAN} is insufficient to support our multi-scale self-watermarking design for tampering recovery.

To encode image content into a recovery watermark, the total number of hiding bits $|h|$ must satisfy the capacity constraint introduced in \cref{eq:capacity_constraint}.  
Thus, instead of embedding all single-scale discrete latent tokens $(x_1, x_2, \ldots, x_T)$, we must reduce their total bit-length.  
Under a single-scale quantization model, two strategies are possible—\emph{partial token selection} or \emph{spatial downsampling}—but both introduce critical limitations:

\textbf{(1) Partial Token Selection.} 
Selecting only a subset of discrete latent tokens $(x_1, x_2, \ldots, x_t)$ with $t < T$ reduces the bit-length to fit within the available watermark capacity $|m|$.  
However, these tokens represent only a spatially limited region of the original image, leading to incomplete recovery.  
Such partial information is unreliable for reconstructing large tampered regions, and becomes unusable when the missing or altered area lies outside the selected token subset.
This strategy is similar to the multiple-scale quantization we adopted in \cref{sec:quantization_watermarking}, but is not feasible to single-scale quantization.
While multi-scale quantization naturally provides coarse-to-fine representations at different resolutions, single-scale quantization lacks this structure, making partial token selection fundamentally unsuitable.

\textbf{(2) Spatial Downsampling.} 
An alternative is to downsample the input image prior to quantization, which reduces the number of tokens $T$ and therefore the required bits.
Yet, as illustrated in \cref{fig:vqgan_explain}, downsampling removes substantial details—including textures, object boundaries, and fine structures such as the bird’s feather contour—causing significant degradation in the recovered image quality.  
This loss is inherent to single-scale quantization and cannot be compensated by upsampling or post-processing.

\begin{table}[h]
    \caption{Comparison of image quality across different recovery approaches. 
    All deepfake images were generated from the ImageNet-S dataset using Stable Diffusion.
    }
    \centering\setlength{\tabcolsep}{4.5pt}\scalebox{0.70}{
        \begin{tabular}{l ccc}
            \toprule
            Methods & SSIM ($\uparrow$) & PSNR ($\uparrow$) & LPIPS ($\downarrow$)\\                        
            \midrule
            HiNet \cite{jing2021hinet}          & 0.7024 & 18.7884 & 0.0375\\
            RePaint \cite{lugmayr2022repaint}   & 0.7832 & 18.9624 & 0.0436\\
            
            \midrule    
            VQGAN \cite{esser2021VQGAN} ($V$ = 1024)        & 0.7987 & 22.6042 & 0.0208\\
            VQGAN \cite{esser2021VQGAN} ($V$ = 16384)         & 0.8029 & 23.1354 & 0.0199\\
            VAR \cite{tian2024VAR}              & 0.7993 & 16.5914 & 0.0249\\
            \midrule  
            \textbf{hidden code (Ours)}  & \textbf{0.8238} & \textbf{24.6575} & \textbf{0.0190}\\
            \textbf{hidden code + Transformer (Ours)}  & 0.8189 & 23.7612 & 0.0198\\
            \textbf{hidden code + conditional Transformer  (Ours)}  & 0.8156 & 23.5550 & 0.0203 \\
            \bottomrule
        \end{tabular}}
    
    \label{tab:exp_structure_metric}
\end{table}

\section{Ablation Studies}\label{Sec: More Results}
\subsection{Image Quality}
The quantitative image quality results corresponding to the factual retrieval results in \Cref{tab:exp_retrieval} are shown in \Cref{tab:exp_structure_metric}. 
Our method achieves consistently higher SSIM and PSNR than HiNet and RePaint, indicating better structural and visual recovery, and is slightly better than VQGAN-based approaches, with overall metric values remaining moderate.
This behavior can be attributed to the information loss introduced by the quantization process described in \cref{eq:quantization}, where each image patch is mapped to its nearest codebook vector. 
Consequently, the quantized image preserves perceptual similarity to the original but deviates at the pixel level.

\subsection{Content-Dependent Watermark (CDW) for Preventing Forgery}
\label{sec:cdw_prevent_forgery}
Beyond deepfake detection, watermark forgery has emerged as another threat that can lead to serious privacy concerns. 
A malicious user may attempt to extract the watermark from a watermarked image and apply it to a clean image, thereby fabricating a counterfeit ``watermarked'' image. 
This vulnerability commonly arises due to the independency between the hidden watermark and the corresponding cover images.
To mitigate this issue, our method embeds content-related information directly into the watermark, forming a Content-Dependent Watermark (CDW). 
Since the specifically designed watermark is tied to the intrinsic image content, transferring it to another image results in inconsistencies, effectively preventing forgery.

\paragraph{Forgery Evaluation Protocol.}
We evaluate watermark forgery resistance under two recent forgery methods, using \textbf{Gaussian Shading \cite{yang2024gaussianshading}} as a representative non-CDW baseline, which embeds the same hidden code but \emph{does not} incorporate our CDW design.
The evaluation includes:

\begin{enumerate}
    \item \textbf{Average Attack (Grey-Box / Black-Box).\footnote{[NeurIPS 2024] Can Simple Averaging Defeat Modern Watermarks?
}} 
    An adversary averages multiple watermarked images to estimate the embedded code, then attempts to apply the estimated watermark to another clean images.  
    The \emph{grey-box} setting assumes access to clean images from the same distribution, while the \emph{black-box} setting uses images from a different dataset.

    \item \textbf{Black-Box Forgery via Noise Estimation \cite{muller2025black}.}
    The attacker use a proxy model to estimate the diffusion initial noise $\tilde{x}_T$ corresponding to a watermarked image $I_w$. 
    Given any clean image $I$, the attacker optimizes the DDIM inversion trajectory such that it aligns with $\tilde{x}_T$, producing a forged image that has the same watermark.
\end{enumerate}
We measure attack success using bit accuracy (BA), where $\text{BA} \approx 0.5$ indicates failure (i.e., random guessing).  
All evaluations were conducted on 1000 sampled images, and we report the average BA.

Across all settings, as shown in Table \ref{tab:exp_forgery}, the non-CDW baseline remains highly vulnerable to forgery attacks, whereas our CDW consistently pushes the attacker’s BA toward 0.5, demonstrating substantially stronger resistance to watermark transplantation and reconstruction.

\begin{table}[h]
    \caption{
    \textbf{Comparison of Forgery Attack Success Rate (Bit Accuracy, BA) under Black-box and Grey-box Settings.}
        We evaluate the success of forgery attacks ($\text{BA} \uparrow$) against a baseline method and against our proposed framework ($\text{BA} \downarrow$).
        The BA quantifies the attacker's ability to guess the hidden code bit, where $\text{BA} \approx 0.5$ signifies a failed attack (random guessing).
        The \textbf{Average Attack} uses $n$ watermarked images to estimate the code.
        The \textbf{Grey-box} setting assumes the attacker has access to the average *clean* image, while the \textbf{Black-box} setting uses an average image from a different dataset.
        Our method consistently forces the attacker's BA toward 0.5, demonstrating a significantly higher level of security compared to the baseline.
    }
    \centering\setlength{\tabcolsep}{3.7pt}\scalebox{0.65}{
        \begin{tabular}{l ccc}
            \toprule
            Methods                          & Gaussian Shading        & Guassian Shading + ours\\
                                             & Forgery BA ($\uparrow$) & Forgery BA  ($\downarrow$)\\                        
            \midrule
            Average attack   ($n=5$, grey box setting)        &  0.8999    & 0.5142 \\
            Average attack   ($n=5$, black box setting)       &  0.8000    & 0.5079 \\
            Average attack   ($n=10$, grey box setting)       &  0.9126    & 0.5125 \\
            Average attack   ($n=10$, black box setting)      &  0.8141    & 0.5047 \\
            Average attack   ($n=20$, grey box setting)       &  0.9328    & 0.5119 \\
            Average attack   ($n=20$, black box setting)      &  0.8564    & 0.5086 \\
            Average attack   ($n=50$, grey box setting)       &  0.9501    & 0.5091 \\
            Average attack   ($n=50$, black box setting)      &  0.8921    & 0.5061 \\
            Average attack   ($n=100$, grey box setting)      &  0.9566    & 0.5089 \\
            Average attack   ($n=100$, black box setting)     &  0.9162    & 0.5082 \\
            Average attack   ($n=200$, grey box setting)      &  0.9569    & 0.5109 \\
            Average attack   ($n=200$, black box setting)     &  0.9179    & 0.5097 \\
            Average attack   ($n=500$, grey box setting)      &  0.9565    & 0.5096 \\
            Average attack   ($n=500$, black box setting)     &  0.9155    & 0.5062 \\
            Black-box Forgery \cite{muller2025black}          & 0.9971     & 0.4962\\            
            \bottomrule
        \end{tabular}}    
    \label{tab:exp_forgery}
\end{table}

\begin{table*}[ht]
    \caption{\textbf{CLIP Similarity Analysis (Semantic Consistency).} We report the CLIP score between the reconstructed image (derived from the embedded hidden code) and the image variants to evaluate both \textbf{robustness} (higher scores for image degradations) and \textbf{forgery resistance} (lower scores for unrelated images). CLIP score ($\uparrow$) indicates better robustness, while CLIP score ($\downarrow$) for the ``Different image'' column indicates better forgery resistance.
    }
    \centering\setlength{\tabcolsep}{4.5pt}\scalebox{0.80}{
        \begin{tabular}{l cccccc|c}
            \toprule
            ImageNet &\multicolumn{6}{c}{CLIP score($\uparrow$)} & CLIP score($\downarrow$)\\            
            Methods {\textbackslash} Metric & Original & Crop & Color jitter & JPEG & Gaussian Blur & Gaussian Noise & Different image \\
                                            &      &$50\%$ & $30\%$ & QF $70\%$ & $5\times 5$ & $\sigma=5$ &  \\
            \midrule  
            VQGAN \cite{esser2021VQGAN} (codebook size $V$ = 1024)  & 0.7939 & 0.7984 & 0.7906 & 0.7730  & 0.8307 & 0.7817 & 0.6787 \\
            VQGAN \cite{esser2021VQGAN} (codebook size $V$ = 16384) & 0.8208 & 0.8229 & 0.8061 & 0.7949  & 0.8572 & 0.8040 & 0.6571\\
            VAR \cite{tian2024VAR}                      & 0.8845 & 0.8692 & 0.8385 & 0.8418  & 0.8582 & 0.8529 & 0.6552\\            
            \midrule           
            \textbf{hidden code (Ours)}                 & 0.9021 & 0.8885 & 0.8503 & 0.8765  & 0.8784 & 0.8948 & 0.6176\\
            \textbf{hidden code + Transformer (Ours)}   & \textbf{0.9061} & \textbf{0.8934} & \textbf{0.8505} & \textbf{0.8766}  & \textbf{0.8821} & \textbf{0.8993} & \textbf{0.6163}\\
            \bottomrule
        \end{tabular}}
    \label{tab:exp_clip_cdw}
\end{table*}

\subsection{CLIP Similarity as an Additional Metric for Detecting Watermark Forgery}
In addition to the bit-accuracy (BA) metric discussed in \cref{sec:cdw_prevent_forgery}, we introduce a complementary evaluation based on CLIP similarity to further assess the forgery resistance of our CDW watermark.  

Since our method embeds content-related hidden codes $h$ into an image $I$ to produce the watermarked image $I_w$, any attempt to transplant the hidden code $h$ into another image $I'$ should result in a \textbf{semantic mismatch}.  
Thus, CLIP similarity serves as a natural metric: if a forged image $I'$ does not align semantically with the embedded code $h$, the CLIP score between the two should drop significantly.

\paragraph{Evaluation Protocol.}
We compute CLIP similarity between the reconstructed image derived from the hidden code and several variants:
\begin{enumerate}
    \item The original image from ImageNet.
    \item Degraded versions of the same image, including:
    \begin{itemize}
        \item 50\% center crop,
        \item Color jitter with 30\% variation (i.e., brightness, contrast, saturation, and hue are each increased or decreased by up to 30\%),
        \item JPEG compression with quality factor = 70,
        \item Gaussian blur with a \textbf{$5 \times 5$} kernel,
        \item Gaussian noise with standard deviation = 5.
    \end{itemize}
    \item Different image $I'$, representing an attempted watermark forgery.
\end{enumerate}
For the ``Different image'' setting, we take the original image $I$ from ImageNet and sample a different, unrelated image $I'$ \emph{also} from ImageNet.
This setup preserves dataset consistency while still evaluating whether the hidden code $h$ can be incorrectly matched to an image with mismatched content.
As shown in Table \ref{tab:exp_clip_cdw}, 
\textbf{high similarity} for 1. and 2. reflects robustness to image variations, while \textbf{low similarity} for 3. demonstrates that hidden codes cannot be meaningfully transferred to unrelated images, confirming ``forgery resistance.''

\end{document}